\documentclass[journal]{IEEEtran}

\ifCLASSINFOpdf
\else
\fi

\usepackage{graphicx}
\usepackage{amsmath}
\usepackage{amssymb}
\usepackage{algorithm}
\usepackage{algorithmic}
\usepackage{multirow}
\usepackage{cite}
\graphicspath{{../figures/}}

\def\A{{\bf A}}

\def\b{{\bf b}}

\def\K{{\bf K}}

\def\k{{\bf k}}

\def\I{{\bf I}}

\def\X{{\bf X}}

\def\P{{\bf P}}

\def\x{{\bf x}}

\def\z{{\bf z}}
\def\Z{{\bf Z}}

\def\W{{\bf W}}
\def\w{{\bf w}}
\def\0{{\bf 0}}
\def\1{{\bf 1}}

\def \DM{{\mathcal D}}
\def \LM{{\mathcal L}}

\def\KM{{\mathcal K}}

\def\XM{{\mathcal X}}

\def\NM{{\mathcal N}}

\def \ZM {{\mathcal Z}}
\def\RB{{\mathbb R}}

\def\Si{\mbox{\boldmath$\Sigma$\unboldmath}}

\def\ie{\emph{i.e. }}

\begin{document}

\title{Zero-Shot Learning with Generative Latent Prototype Model}
\author{Yanan~Li,~\IEEEmembership{Student Member,~IEEE,}
Donghui~Wang,~\IEEEmembership{Member,~IEEE}% <-this % stops a space
\thanks{This work was supported by the National Natural Science Foundation of China
under Grants 61473256 and CKCEST project.}
\thanks{The authors are with the College
of Computer Science, Zhejiang University, Hangzhou,
310027, China e-mail: (ynli@zju.edu.cn, dhwang@zju.edu.cn).}}% <-this % stops a space
%\thanks{This work has been submitted to the IEEE for possible publication. Copyright may be transferred without notice,
%after which this version may no longer be accessible.}}
%\thanks{Color versions of one or more of the figures in this paper are available online at\emph{ http://ieeexplore.ieee.org.}}% <-this % stops a space
%\thanks{Manuscript received April 19, 2016; revised August 26, 2016.}}

%% The paper headers
%\markboth{Journal of \LaTeX\ Class Files,~Vol.~14, No.~8, August~2016}%
%{Shell \MakeLowercase{\textit{et al.}}: Bare Demo of IEEEtran.cls for IEEE Journals}
% The paper headers
%\markboth{Journal of \LaTeX\ Class Files,~Vol.~14, No.~8, August~2015}%
%{Shell \MakeLowercase{\textit{et al.}}: Bare Demo of IEEEtran.cls for IEEE Journals}

% make the title area
\maketitle

\begin{abstract}
Zero-shot learning, which studies the problem of object classification for categories for which we have no training examples, is gaining increasing attention from community. Most existing ZSL methods exploit deterministic transfer learning via an in-between semantic embedding space. In this paper, we try to attack this problem from a generative probabilistic modelling perspective. We assume for any category, the observed  representation, e.g. images or texts, is developed from a unique prototype in a latent space, in which the semantic relationship among prototypes is encoded via linear reconstruction. Taking advantage of this assumption, virtual instances of unseen classes can be generated from the corresponding prototype, giving rise to a novel ZSL model which can alleviate the domain shift problem existing in the way of direct transfer learning. Extensive experiments on three benchmark datasets show our proposed model can achieve state-of-the-art results.
\end{abstract}

%% Note that keywords are not normally used for peerreview papers.
%\begin{IEEEkeywords}
%Zero-shot Learning, transfer learning, probabilistic model.
%\end{IEEEkeywords}
% For peer review papers, you can put extra information on the cover
% page as needed:
% \ifCLASSOPTIONpeerreview
% \begin{center} \bfseries EDICS Category: 3-BBND \end{center}
% \fi
%
% For peerreview papers, this IEEEtran command inserts a page break and
% creates the second title. It will be ignored for other modes.
\IEEEpeerreviewmaketitle

\section{INTRODUCTION}
It has been estimated that humans can easily distinguish between approximately 30000 basic object categories \cite{biederman1987recognition} and many more subordinate ones, such as different species of birds. Without seeing them, human beings can even recognize new unseen categories by leveraging other information (e.g. by reading text descriptions about object categories on the internet). In contrast, encumbered with a lack of adequate data, generally machines can only recognize hundreds or thousands categories. To free recognition tasks from exuberant collecting of large labelled image datasets, zero-shot learning (ZSL) is gaining increasing attention in recent years, which aims to recognize instances from the new unseen categories which have no instances during training \cite{palatucci2009zero, lampert2009learning, socher2013zero}.
With the label sets between seen and unseen categories being disjoint, the key in the general methodology of ZSL is to establish the inter-class connections via intermediate semantic representations,  either manually defined by human experts annotated attributes \cite{farhadi2009describing,parikh2011interactively,mensink2012metric,lampert2014attribute,al2015transfer}, or automatically extracted from auxiliary text sources \cite{berg2010automatic,rohrbach2010helps, norouzi2013zero,rohrbach2013transfer, frome2013devise,akata2016multi}. Unseen categories can thus be predicted by transferring information from the training dataset. As a valuable knowledge base given in advance, in theoretical, the semantic representations of unseen categories are encouraged to be leveraged in any stage during ZSL. However, most recent works mainly focus on exploring these representations to construct a more effective classifier during testing. While, how to explore them during training to learn more generalized is equally important but still left far from being solved, since the quality of semantic representation predictor is much more rewarding \cite{gan2016learning}. In addition, due to the disjoint data distribution between seen and unseen classes, direct knowledge transfer will cause the domain shift problem during ZSL, leading to degraded performance.
\begin{figure}
  \centering
  % Requires \usepackage{graphicx}
  \includegraphics[width=0.35\textwidth]{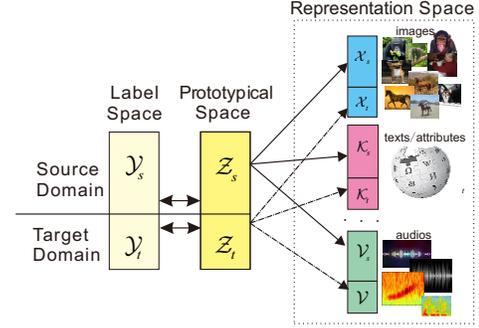}\\
  \caption{Illustration of the proposed probabilistic generation process.}
  \label{fig:zsl}
\end{figure}

In this letter, we tackle these challenges with ideas from generative learning. We posit that there exists a latent space, as illustrated in Fig.\ref{fig:zsl}, where each object category is encoded by a unique data (called prototype) essentially. Any type of object representations, e.g. texts or images, are generated from its corresponding prototypes from different perspectives. This supposition is inspired from the cognitive process of human beings, who have remarkably ability of generating various representations, e.g. images, audios, texts, from high-level category labels 
\cite{coyne2001wordseye,lake2015human,rezende2016one}. For example, it is almost effortless for people to imagine the different picture/audios, given the label 'penguin' and 'sparrow'. 

To mathematically formulate this institution, we assume that the latent prototypes obey a prior distribution where one can draw samples from. During the data generation process, the category prototype is first generated, from which then different observed representations can be developed. Based on this generation process, we further explore the semantic representations given beforehand to make the training process generalize well across unseen categories. A simple ZSL method encompassing different strategies is proposed to solve the domain shift problem by generating virtual unseen instances. Ahead of time, we further give a simple property about these semantic representations as a basic condition for their application in ZSL. Extensive experiments on real world datasets show our proposed method can achieve state-of-the-art results.

%In the proposed model, we assume there exists a latent prototypical space, which contains the prototypes of object categories and encodes the true semantic relationship among them. Each prototype abstracts the common essential characteristics of different representations. Assuming that latent prototypes obey a prior distribution where one can draw samples from, the category prototype is first generated. Then from it, different observed representations can be developed, as is shown in Fig.\ref{fig:zsl}. Actually, each type of representations depicts categories from different perspectives.
%Based on this generation process, we propose a simple ZSL method to solve the domain shift problem by generating virtual instances for target domain in the training stage.
%
%Meanwhile, we further give a simple property about these observed representations as a basic condition for their application in ZSL. Extensive experiments on real world datasets show our proposed method can achieve state-of-the-art results.

\section{GENERATIVE LATENT PROTOTYPE MODEL}

\subsection{Problem Setting}
Following convention, let $\LM_s = \{y_s^1, ..., y_s^k\}$ and $\LM_t = \{y_t^1, ..., y_t^l\}$, $\LM_s \bigcap \LM_t = \emptyset$ be disjoint label sets of seen and unseen classes in the source domain $\DM_s$ and target domain $\DM_t$, respectively. Each category corresponds one-to-one to a unique prototype in the latent space $\ZM$, denoted as $\ZM_s = \{\z_s^1, ..., \z_s^k\}$ and $\ZM_t = \{\k_t^1, ..., \k_t^l\}$. We assume there are two different types of observed category representations in ZSL, i.e. visual features $\XM$ and semantic features $\KM$ (defined by attributes/texts). In $\KM$, all categories in $\LM_s$ and $\LM_t$ are embedded as $\KM_s$ and $\KM_t$ in advance. 
Given a new test image feature $\x_t$, the task of zero-shot learning is to construct a classifier $f: \arg\max_{l} \log p(y_t^l|\x_t)$ by making use of image source dataset $\{\x_i, y_i\} \subset \XM \times \LM_s$ and all available information in $\KM$.
%
%In the training stage, we have only visual examples for $\LM_s$, denoted as $\XM_s$, while examples for $\LM_t$ are unseen. For semantic textual features $\KM$, representations for both $\LM_s$ and $\LM_t$ are available, denoted by $\KM_s$ and $\KM_t$. The task of learning with disjoint source and target classes is to construct a classifier: $f: \arg\max_{l} \log p(y_t^l|\x_t)$ by making use of source examples $\{(\x_i, \k_i, y_i)\}_{i=1}^N \subset \XM_s \times \KM_s \times \LM_s$ and target category representations $\KM_t$.

\subsection{ZSL with A Latent Prototypical Space}
\label{sec:gen}
Let us motive our approach from a generative probabilistic modelling perspective.
%This will in turn provide a basis for structuring our learning method for zero-shot recognition.
We assume that for each category, there are several different observed representations, e.g. images, texts or audio. Each describes the category from a specific perspective and is developed from the category prototype (an original or first model of something from which other forms are copied or developed\footnote{Definition taken from Merriam-Webster.com dictionary.}), which abstracts the common essence about the category from these representations. In the latent prototypical space, we assume the manifold structure of prototypes encodes the underlying semantic relationship between different categories. The similar assumption has been successfully applied in \cite{wang2016relational,fu2015transductive,li2015semi,zhang2015zero} .

To formulate this process, we use the random variable $\z \in \RB^m$ to denote the category prototype in the latent space $\ZM$ and $\k \in \RB^d$ to denote the observed representation for clarity. For each observation, the generation process, shown in Fig.1, is as follows.
\begin{itemize}
  \item Choose a category prototype $\z_c \sim p(\z_c)=Cat(\Z)$, where $Cat(\Z)$ is the categorical distribution and $\Z = [\z_1, ..., \z_{k+l}]$ contains all the prototypes.
  \item Generate observations of the category prototype $\z_c$ as $\k \sim p(\k|\z_c)$. Without loss of generality, we assume $p(\k|\z_c)$ is a linear Gaussian distribution.
\end{itemize}

Specially, in ZSL, we have two types of representations, i.e. image $\x$ and text $\k$. For  class $c$, we assume $\x \sim p(\x|\z_c) = \NM(\x|\P_x \z_c, \Si_x)$ and $\k \sim p(\k|\z_c) = \NM(\P_k \z_c, \Si_k)$ for $\XM$ and $\KM$, respectively. $\P_x \in \RB^{d\times m}$ and $\P_k \in \RB^{a \times m}$ are the projection matrices, $\Si_x \in \RB^{d\times d}$, $\Si_k \in \RB^{a\times a}$ are their covariance matrices. Because $\x$ and $\k$ depict different aspects of the category, we give the conditional independence for them, i.e. $p(\x, \k |\z_c) = p(\x|\z_c) p(\k|\z_c)$.

Given an instance $\x_t$ from $\DM_t$, its class label is predicted as
\begin{equation}
  y_t = \arg\max_l p(l|\x), l \in \LM_t.
  \label{eq:zsl}
\end{equation}

To aid the prediction in Eq.\ref{eq:zsl}, we need to use the textual category representation $\k$ as an in-between layer to decouple images $\x$ from label $l$, due to its easy accessability and semantic integrity.
%To aid the prediction in Eq.\ref{eq:zsl}, we need to use the representation of $\k$ as an in-between layer to decouple the layer of $\x$ from the layer of $l$.
Common practice is to assign $\k$ on a per-class basis, or a per-image basis. The former is particularly helpful, since it allows the minimum effort of annotating a theoretically unlimited number of unseen categories. For convenience, we consider the former case to give our method, which can easily be extended to the latter.

The per-class annotation allows a deterministic labelling of the intermediate semantic layer $\k$. Therefore, the label prediction in Eq.\ref{eq:zsl} becomes:
\begin{equation}
  y_t= \arg\max_l p(\k_t^l | \x), l \in \LM_t,
\label{eq:zslk}
\end{equation}

\subsection{Prerequisite Condition}
Before introducing the proposed method, we first give a discussion about the category representation $\KM$ to assist the ZSL task. Obviously, not any representation has the ability of transcending class boundaries and be used to transfer knowledge for making predictions. It should meet the following property.

\textbf{Basic Property.} \emph{For $\exists \k_t^i \in \KM_t$, if $\k_t^i \not\in range(\K_s)$, where $\K_s = [\k_s^1, ..., \k_s^k]$ and $range(\K_s)$  denotes the column space of the matrix $\K_s$, then  $\KM=\{\KM_s, \KM_t\}$ has no transferability for ZSL.}

\emph{Proof.} For $\k_t^i$, if $\k_t^i \not\in range(\K_s)$, $\forall \alpha \in \RB^{k}$, $\k_t^i \perp \K_s \alpha$, i.e. $\k_t^i$ is not in the subspace spanned by all seen classes. Given $\x_t$, $\forall \k_t^l \in \K_t$, $p(\k_t^l|\x_t)$ has the same possibility. Thus Eq.\ref{eq:zslk} can't make predictions.

This property describes a kind of criterion to  evaluate whether a specific category representation is transferrable intuitionally. Similar conclusion about binary attribute representations has been discussed in \cite{palatucci2009zero,romera2015embarrassingly}.

\subsection{Generative Latent Prototype Model (GLaP)}
Based on the above discussion, we propose the solution for our probabilistic model in Eq.\ref{eq:zslk}.
\subsubsection{Learn directly from $\DM_s$}
%Due to the absence of target domain instances during training, one natural solution for Eq.\ref{eq:zslk} is to replace $p(\k_t^c|\x)$ directly with what is learned from training data in $\DM_s$, i.e.
Due to the absence of target domain instances during training, one natural solution for Eq.\ref{eq:zslk} is to learn $p(\k|\x)$ directly from training data in $\DM_s$ by maximizing its log likelihood, i.e. $\max \log_{\DM_s} p(\k|\x)$, where
%by optimizing $\arg\max_{\A, \b} \log_{\DM_s}p(\k|\x)$, where
\begin{equation}
\begin{split}
 % p(\k|\x) & = \int p(\k, \z | \x)d\z \\
%   p(\k|\x)  \propto\int p(\k, \x|\z_c)p(\z_c)d\z_c =\int p(\k|\z_c)p(\x|\z_c)p(\z_c)d\z_c\\
   p(\k|\x)  \propto \int p(\k|\z_c)p(\x|\z_c)p(\z_c)d\z_c
  %& = \int p(\k|\z)p(\x_s|\z)p(\z)d\z, \\
  % &= \int \NM(\k|\P_k \z_c, \Si_k) \NM(\x|\P_x \z_c, \Si_x) Cat(\z_c|\Z) d\z_c,
\end{split}
\label{eq:zslt}
\end{equation}
All three distributions are in the exponential family. $p(\k|\x)$ is actually a linear Gaussian distribution, i.e. $\NM(\A \x + \b, \Si)$, where $\A \in \RB^{a \times d}$ and $\b \in \RB^a$ establish the connection between $\x$ and $\k$ and can be solved in closed form \cite{bishop2007pattern}. Take the simplest case for example. When $\Si = \I, \b=\0$, $\A = \K\X^T(\X\X^T)^{-1}$, where $\K = [\k_1, ..., \k_N]$ and $\X = [\x_1, ..., \x_N]$.

However, due to $\LM_s \bigcap \LM_t = \emptyset$, the underlying data distributions of the object categories differ. Approximating the ideal function $p(\k | \x)$ for $\DM_t$ using Eq.\ref{eq:zslt} suffers from a domains shift problem \cite{fu2015transductive}. On one hand, it just optimize the source domain where labelled information of target classes is missing. On the other hand, $\XM$ and $\KM$ may differ in the semantic relationship among different classes, due to their emphasis in the generation process. Therefore, using Eq.\ref{eq:zslt} without any adaptation to the target domain will cause significant performance degradation \cite{wang2016relational}. One natural solution to this problem is  loading a small amount of instances for target classes in the training stage to adjust Eq.\ref{eq:zslt}.

\subsubsection{Learn from a virtual $\tilde{\DM}_t$ }
In the above discussion, we assume the prototypical space $\ZM$ encodes the essence information of categories and also the true semantic relationship among different categories. Let us denote by $\Z_s = [\z_s^1, ..., \z_s^k]$ and $\Z_t = [\z_t^1, ..., \z_t^l]$ the prototypes of all classes in $\DM_s$ and $\DM_t$, respectively. And $\W =[\w_1, ..., \w_l] \in \RB^{k\times l}$ encodes their semantic relationships. Instead of the graph-based relationship \cite{rohrbach2013transfer,fu2015transductive,fu2015zero}, $\W$ is constrained to be linear in this letter, i.e. $\forall \z_t^i$, $\z_t^i = \Z_s \w_i$. Based on our generation process, given the prototype $\z_t^i$, $\x$ in this class  is actually a Gaussian distribution, i.e.
\begin{equation}
\begin{split}
p(\x|\z_t^i) = \NM(\P_x \z_t^i, \Si_x) = \NM(\P_x \Z_s \w_i, \Si_x)
\end{split}
\label{eq:xt}
\end{equation}
where $\w$ encodes its relationship with $\Z_s$. Thus, to generate unseen instances, we need to estimate the two parameters $\P_x$ and the invisible $\Z_s$ and $\w_i$ from training data.

First, we estimate $\w_i$ by means of textual representation $\KM$, by the following function:
\begin{equation}
  %\arg \min_{\W} ||\K_t - \K_s \W||^2_F + \Omega(\W),
  \w_i = \arg \min_{\w_i} ||\k_t^i - \K_s \w_i||^2_F + \Omega(\w_i),
  \label{eq:w}
\end{equation}
where $\Omega$ is the regularizer of $\w_i$, common choice is $\ell_2$ or $\ell_1$ norm \cite{bofill2001underdetermined,sprechmann2010dictionary,cheng2013sparse}. While in the per-image basis, its relationship $\w_i$ can be obtained using the mean representation.

Second, we further simplify Eq.\ref{eq:xt} by estimating $\P_x \Z_s$ instead of the explicit computation of $\Z_s$ and $\P_x$ separately. Considering the generation process in Sec.\ref{sec:gen}, we assume different types of representations are produced independently. Given $\DM_s$, we obtain the prototype for each seen class by maximizing the likelihood of visual representations, i.e.
\begin{equation}
  \arg\max_{\P_x\z_s^i} \prod_{j=1}^{N_s^i} p(\x_j), ~ p(\x_j) = \int p(\x_j|\z_s^i) p(\z_s^i) d\z_s^i,
  \label{eq:pz}
\end{equation}
where $N_s^i$ denotes the number of training examples in the class $\z_s^i$. Optimizing Eq.\ref{eq:pz} gives rise to $\P_x \z_s^i = \frac{1}{N_s^i}\sum_{j=1}^{N_s^i} \x_j, \x_j \in \{\x|y=i\}$, which is the mean vector. We denote it as $\bar{\x}_s^i$ for clarity. Substituting parameters in Eq.\ref{eq:xt}, we obtain:
\begin{equation}
  p(\x|\z_t^i) \sim \NM(\bar{\X}_s \w_i, \Si_x)
  \label{eq:dist-xt}
\end{equation}
where $\bar{\X}_s = [\bar{\x}_s^1, ..., \bar{\x}_s^k]$ contains all the mean vectors of source classes in $\DM_s$ and $\Si_x = \sigma^2\I$ is a predefined covariance matrix. From this distribution, a bunch of virtual instances for unseen category can be randomly produced, denoted as $\tilde{\DM}_t = \{(\tilde{\x}_t^i, \tilde{\k}_t^i,  \tilde{l}_i)\}$. Thus, an alternative strategy for ZSL is to learn directly from $\tilde{\DM}_t$.

\subsubsection{Final Objective}
Combining the above two parts, we need the projection $\A$ in Eq.\ref{eq:zslt} on the one hand to be optimal for $\DM_s$, on the other hand to be optimized for unseen categories to solve the domain shift problem. We use a trade-off parameter $\lambda$ to adjust these two effects. The overall objective function is:
\begin{equation}
  %\arg\max_{\A} \lambda p(\k|\x) + (1-\lambda) p(\k|\x),
  \arg\max_{\A} \lambda \log_{\DM_s} p(\k|\x) + (1-\lambda) \log_{\tilde{\DM}_t}p(\k|\x),
  \label{eq:obj}
\end{equation}
where $\log_{\DM_s} p(\k|\x)$ means calculating $p(\k|\x)$ from $\DM_s$. This objective function gives us 3 strategies for predicting in ZSL. (1). When $\lambda = 1$, learn directly from $\DM_s$; (2). When $\lambda = 0$, learn from $\tilde{\DM}_t$; (3). When $0<\lambda<1$, learn from both $\DM_s+\tilde{\DM}_t$. Their respective performance is showed in later experiments.

\begin{algorithm}[htp]
  \caption{The proposed algorithm}
  \begin{algorithmic}[1]
  \REQUIRE Semantic representations $\K_s$ and $\K_t$, source data $\{\x_i, \k_i, y_i\}_{i=1}^N, \k_i = \k_s^{y_i}$. 
    \STATE Extract relational knowledge by Eq.\ref{eq:w}.
    \STATE For each unseen class, generate $m$ virtual instances by Eq.\ref{eq:dist-xt}.
    \STATE Learn $\A = (\lambda \X\X^T +(1-\lambda) \tilde{\X}\tilde{\X}^T)^{-1} (\lambda \K \X^T + (1-\lambda)\K\tilde{\X}^T)$.  
    \STATE Predict unseen label by Eq.\ref{eq:zslk}. 
  \end{algorithmic}
\end{algorithm}

% needed in second column of first page if using \IEEEpubid
%\IEEEpubidadjcol

\section{EXPERIMENTS AND RESULT ANALYSIS}
In order to assess the validity of the statements we made, we conducted a set of experiments on three real world datasets.

\subsection{Experimental Setup}
\textbf{Datasets.} We test our work on three benchmark image datasets. Animals with Attributes (\textbf{AwA}) \cite{lampert2009learning} consists of 30,475 images of 50 image classes, each paired with 85 human-labelled attributes. We follow the usual procedure~\cite{lampert2009learning}, \ie 40 classes for training and 10 for testing. Caltech-UCSD Birds-200-2011 (\textbf{CUB}) \cite{wah2011caltech} is a fine-grained dataset with 312 attributes annotated for 200 bird classes. It contains 11,788 images in total. Following \cite{akata2015evaluation}, we use the same 150/50 class split for training and testing. Standford Dogs (\textbf{Dogs}) \cite{khosla2011novel} contains 19,501 images of 120 fine-grained dog species, with no attributes annotated. We use 90 classes for training and the rest for testing.

\textbf{Choices for $\XM$ and $\KM$.} We mainly use two different types of observed representations in this letter. We use 3 types of deep features for $\XM$, extracted from 3 popular CNN architectures, \ie VGG~\cite{simonyan2015very}, GoogLeNet~\cite{szegedy2014going} and ResNet~\cite{he2015deep}. We extract respectively 1000D, 1024D and 1000D features from these CNNs, which are denoted as $fc8$, $goog$ and $res\_fc$. They are both low-dimensional and high-semantic features. For the semantic textual representation $\KM$, 2 different types are used, \ie continuous human-annotated attributes (denoted as $A$) and 3 kinds of word vectors learned automatically from Wikipedia ($skipgram$, $cbow$~\cite{mikolov2013distributed} and $glove$~\cite{pennington2014glove}).

\begin{table*}[!htp]
  \centering
%  \begin{threeparttable}[b]
    \caption{Accuracy (\%) on AwA, CUB and Dogs.`+' is the concatenation of two features. `--' means no results reported.}
    \begin{tabular}{|c|c|c|c|c|c|c|c|c|c|c|}
      \hline
      \multicolumn{2}{|c|}{} & \multicolumn{3}{c|}{A} & \multicolumn{3}{c|}{W} & \multicolumn{3}{c|}{A+W}\\
      \hline
      Datasets & Feature & Baseline & \textbf{GLaP \#1} & \textbf{GLaP \#2} & Baseline & \textbf{GLaP \#1} & \textbf{GLaP \#2} & Baseline & \textbf{GLaP \#1} & \textbf{GLaP \#2}\\
      \hline
      \multirow{2}{*}{AwA} & \emph{goog} & 65.91 & 72.52& 76.63& 55.86& 73.90& 64.68& 73.68 & \textbf{81.29} &80.84 \\
      \cline{2-11}
                                      & \emph{rec-fc}  &74.73 & 68.54 &  71.36& 58.23& 70.10 &  61.84& 77.56 & 74.85 &\textbf{80.24}\\
      \hline
      \multirow{2}{*}{CUB} & \emph{goog} &32.24 &42.45& 32.79 & 22.24&19.62 & 25.76 &42.14 &46.83 & \textbf{50.38} \\
      \cline{2-11}
                                      &\emph{ rec-fc } & 30.66& 39.28 & 32.48&21.21 &18.59 &  24.76& 42.97&42.14 & \textbf{47.24}\\
      \hline
      \multirow{2}{*}{Dogs} & \emph{goog} & -&- & - &20.49 & 27.77 &28.61 &- &- & -\\
      \cline{2-11}
                                      &\emph{ rec-fc}  & -& -& - & 23.59&23.17 & \textbf{30.52} &- &- & -\\
      \hline
    \end{tabular}
    \label{tab:prop}
\end{table*}

\subsection{Evaluation on the strategy of loading testing instances}
In the first set of experiments, we test the performance of our proposed method under various image features $\XM$ and textural descriptions $\KM$. We consider 3 different kinds of strategies corresponding to diverse $\lambda$ in Eq.\ref{eq:obj}.
%When $\lambda = 1$, we learn only from $\DM_s$, which is exactly the baseline. When $\lambda = 0$, we generate a small bunch of data and learn from $\tilde{\DM}_t$, denoted as $\textbf{Ours \#1}$.
They are \textbf{Baseline} (\emph{Learn from $\DM_s$ when $\lambda = 1$}), \textbf{GLaP \#1} (\emph{Learn from $\tilde{\DM}_t$, when $\lambda$=0}) and \textbf{GLaP \#2} (\emph{Learn jointly from $\DM_s$ and $\tilde{\DM}_t$, when $\lambda = \frac{1}{2}$}). In the latter two methods, we generate a small bunch of data from Eq.\ref{eq:dist-xt}. For clarity, we use only two types of $\XM$, i.e. $goog$ and $rec\_fc$ and three different types of $\KM$, i.e. manual attributes ($A$), word vectors ($W$) and a concatenation of attributes and word vectors ($A+W$). Experimental results are shown in Tab.\ref{tab:prop}.

Comparing the results of  \textbf{GLaP \#2} with baseline, we find loading virtual testing instances during training can boost  the ZSL performance greatly, regardless of different category representations.  On AwA dataset, it achieves the highest $80.84\%$, $7\%$ more than our baseline. Similarly, the same degree of improvement can be observed on CUB and Dogs, although the performance is not as good as in AwA. On CUB, the reason of performance degradation is that the much finer granularity can hardly be reached by these general deep features, which lower the discrimination ability. In addition, when adopting just one type of category representations, i.e. either $A$ or $W$, the ZSL performance is a little decreased than in the case of $A+W$, with the largest difference being almost $25\%$ in CUB. Indirectly, this phenomenon proves our standpoint in the proposed probabilistic model, i.e. different types of observed representations describe the category from different perspectives. To some extent, they can provide complimentary information about categories to improve recognition performance.

\begin{figure}[!htp]
  \centering
  \includegraphics[width=0.4\textwidth]{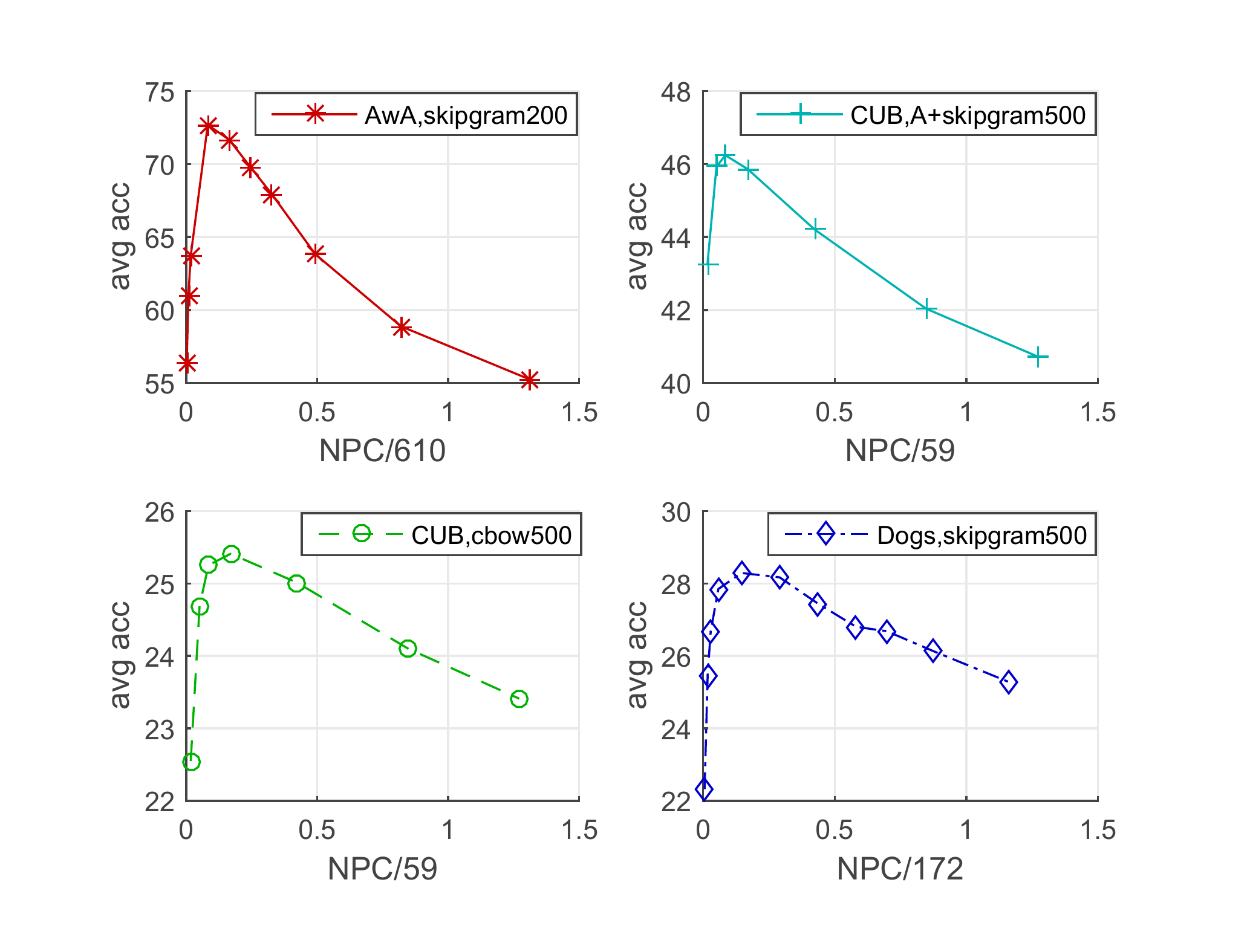}\\
  \caption{Accuracy improvement over baselines by using different number of virtual instances (NPC). Results are obtained using $goog$. }
  \label{fig:npc}
\end{figure}

Generated virtual instances $\tilde{\DM}_t$  alone can be used to solve ZSL problem in Eq.\ref{eq:obj} as well. Comparing with baseline, \textbf{GLaP \#1} usually achieves better results. On AwA, using the textual representation, it even obtain the astonishing highest accuracy, $81.29\%$. This result shows the potential of unsupervised-learned word vectors in boosting ZSL performance, while refraining from the cumbersome human work of annotating attributes. In contrast with \textbf{GLaP \#2}, $\tilde{\DM}_t$ alone performs almost equally, in some cases even better, e.g. on AwA. Impact of the size of $\tilde{\DM}_t$ is also shown in Fig.\ref{fig:npc}. With a small number of generated instances, it performs quite well.

Specially, we note that in the generation process of virtual instances, the mean vector of each class, denoted as $\bar{\DM}_s$, plays  an important role. Therefore, we further conduct experiments on the strategy of learning jointly from $\tilde{\DM}_t$ and $\bar{\DM}_s$, denoted as \textbf{GLaP \#3}. The experimental results shown in Tab.\ref{tab:ours3} demonstrate that this strategy can basically achieve as good performance as \textbf{GLaP \#2}. It is worth mentioning that with a little semantic information of $\KM$ and only a few mean vectors in $\XM$, the proposed method works well. We assume this will benefit online zero-shot recognition.

\begin{table}
  \centering
  \caption{Accuracy (\%) achieved by \textbf{GLaP \#3}.}
  \begin{tabular}{|c|c|c|c|c|}
   \hline
   Datasets & Features & A & W & A+W \\
   \hline
   \multirow{2}{*}{AwA} & $goog$ & 70.63 &68.25 & 80.37\\
   \cline{2-5}
                                   & $rec\_fc$ &74.26&69.53 &\textbf{81.50 }\\
   \hline
   \multirow{2}{*}{CUB} & $goog$ & 33.79 & 25.38 & \textbf{50.66} \\
   \cline{2-5}
                                  & $rec\_fc$ &33.55 &25.41 & \textbf{47.41}\\
   \hline
   11
   \multirow{2}{*}{Dogs} & $goog$ & - & 28.95 & -\\
   \cline{2-5}
                                   & $rec\_fc$ &- & 25.34& -\\
   \hline
  \end{tabular}
  \label{tab:ours3}
\end{table}

\subsection{Comparison with the State-of-the-art}
In the third set of experiments, to better show the zero-shot performance of the proposed methods, we use the concatenation of $fc8$, $goog$ and $res\_fc$ for $\XM$ and compare it with several state-of-the-art ZSL methods. They are SJE \cite{akata2015evaluation}, HAP  \cite{huang2015learning}, ZSLwUA \cite{jayaraman2014zero}, PST \cite{rohrbach2013transfer}, TMV\cite{fu2015transductive}, AMP \cite{fu2015zero}, UDA \cite{kodirov2015unsupervised} and UDICA \cite{gan2016learning}. 
%In the experiments, we choose the appropriate pair of $\XM$ and $\KM$ for each dataset, based on the above discussion.

For simple comparison, we use same settings and author-provided results. Results in Table.\ref{tab:exp3} testify the effectiveness of generating virtual testing instances in our methods. They achieve the state-of-the-art results. On CUB and Dogs, our methods even exceeds SJE, whose $\KM$ is constructed with specific word vectors learned from a specific text corpus. One thing should be noted that since our results are obtained based on the baseline method, we expect the result to be further improved when incorporated with other ZSL methods in this table, which is also a work in the future.
%\begin{table}[!htp]
%\centering
%%\begin{threeparttable}[b]
%\caption{Comparisons with state-of-the-art ZSL methods. }
%\begin{tabular}{l l c c c}
%  \hline
%  \hline
%  Methods & $\KM$ & AwA  & CUB & Dogs \\ \hline
%  SJE                          & A/W     & 66.7      & 50.1     & 33.0\\
%  HAP                         & A         & 45.6      & 17.5     & -\\
%  ZSLwUA                  & A          &43.01     &-           & -\\
%  PST                         & A          &42.7       &-           &-\\
%  TMV                       & A+W     &80.5       &47.9      &-\\
%  AMP                       & A+W     &66          &-          &-\\ \hline
%%  baseline                  & A+W     &80.13     &44.62    &23.59\\ \hline
%  \textbf{Baseline}& A+W     &80.13     &44.62    &23.59\\
%  \textbf{Ours \#1} & A+W    &81.29     &46.83    &27.77\\
%  \textbf{Ours \#2} & A+W    &\textbf{83.45}     &52.62    &\textbf{31.93}\\
%  \textbf{Ours \#3} & A+W  &\textbf{83.24}     &\textbf{52.79  }  &29.96\\
%  \hline \hline
%\end{tabular}
%%\begin{tablenotes}
%%  \footnotesize
%%  \item[1] Notations A and W represent attribute knowledge and word vector knowledge, respectively.
%%  \item[2] *: Method uses word vectors learned from a specific corpus. `--' means no result reported.
%%\end{tablenotes}
%%\end{threeparttable}
%\label{tab:exp3}
%\end{table}
\begin{table}[!htp]
\centering
%\begin{threeparttable}[b]
\caption{Comparisons with state-of-the-art ZSL methods. }
\begin{tabular}{l l c c c}
  \hline
  \hline
  Methods & $\KM$ & AwA  & CUB & Dogs \\ \hline
  SJE                          & A/W     & 66.7      & 50.1     & 33.0\\
  HAP                         & A         & 45.6      & 17.5     & -\\
  ZSLwUA                  & A          &43.01     &-           & -\\
  PST                         & A          &42.7       &-           &-\\
  TMV                       & A+W     &80.5       &47.9      &-\\
  AMP                       & A+W     &66          &-          &-\\ \hline
%  baseline                  & A+W     &80.13     &44.62    &23.59\\ \hline
  %\textbf{Baseline}& A+W     &80.13     &44.62    &23.59\\
  \multirow{3}{*}{\textbf{GLaP}} & A    &77.57    &41.79    &-\\
   & W  &72.49     &28.28    &\textbf{31.93}\\
   & A+W  &\textbf{83.45}   &\textbf{52.79}  &-\\
  \hline \hline
\end{tabular}
\label{tab:exp3}
\end{table}

\section{Conclusion}
In this letter, we proposed a generative latent prototype model for zero-shot learning. We assume observed category descriptions are developed from the category prototype, which is able to encode the true semantic relationship among different categories. Based on this assumption, virtual instances for unseen categories in the target domain can be produced, which give rise to the improved efficiency of our ZSL model. Experiments showed it achieved the state-of-the-art results on three benchmark datasets.
%\appendices
%\section{Proof of the First Zonklar Equation}
%Appendix one text goes here.

% you can choose not to have a title for an appendix
% if you want by leaving the argument blank
%\section{}
%Appendix two text goes here.

% use section* for acknowledgment
%\section*{Acknowledgment}

%The authors would like to thank...

% Can use something like this to put references on a page
% by themselves when using endfloat and the captionsoff option.
\ifCLASSOPTIONcaptionsoff
  \newpage
\fi

%\begin{thebibliography}{1}
%
%\bibitem{IEEEhowto:kopka}
%H.~Kopka and P.~W. Daly, \emph{A Guide to \LaTeX}, 3rd~ed.\hskip 1em plus
%  0.5em minus 0.4em\relax Harlow, England: Addison-Wesley, 1999.
%
%\end{thebibliography}

\bibliographystyle{IEEEtran}
\bibliography{spl16}
%% if you will not have a photo at all:
%\begin{IEEEbiographynophoto}{John Doe}
%Biography text here.
%\end{IEEEbiographynophoto}
%
%% insert where needed to balance the two columns on the last page with
%% biographies
%%\newpage

\end{document}